# Using technology to save endangered languages[1]

*How linguists and computer scientists came together in a retreat to explore ways to advance automatic speech recognition for endangered languages*

Hilaria Cruz and Joseph Waring

**Abstract**
In August 2018 a retreat in Quechee, Vermont, brought together computer scientists specializing in natural language processing, linguists, native speakers, and endangered language activists under one roof. During the retreat, participants discussed ways to utilize the latest advances in Automatic Speech Recognition, especially neural networks, to transcribe endangered languages and tackle the difficulties of transcribing natural language, addressing what is known as "the bottle neck" of language transcription. In a relaxed environment where work was mixed with fun, everyone who participated became friends quickly and interacted with collegiality, exhibiting great potential for future collaborations.

## 1.0 Background

Automatic Speech Recognition (ASR) is a burgeoning technology with near limitless potential. Voice recognition, which was once considered science fiction, is now commonplace: Alexa, Siri, and OK Google live on countertops and bedside tables in our homes. Nevertheless, this technology has been limited to major, dominant languages such as English, German, and Spanish.

In the 19th and 20th centuries, nation states sought to stamp out linguistic and cultural diversity, setting language decline in rapid motion all over the world. Most of the world's 7,000 languages will fall out of use by the end of this century if language loss continues at its current rate, limiting humanity's ability "to appreciate the full creative capacities of the human mind" (Mithun 1998:189).

The rapid development of ASR technologies, especially of neural networks, could expedite the transcription, translation, and linguistic annotation of endangered languages, providing resources for research, revitalization, and promotion. Before this can be achieved, linguists and computational scientists must overcome several technological and methodological obstacles.

Traditional ASR models, such as Markov chains, forced alignment, and hidden Markov models, require vast quantities of training data to get a model running (Michaud et al. 2018). Most endangered languages lack large corpora, alphabets, or speakers for data elicitation. A great number of these languages only have a few speakers left. Manual transcription and annotation of speech is time-consuming. As a native speaker of Chatino, it takes me on average 30 minutes to transcribe one minute of text. For non-native speakers, the process could take much longer.

---

[1] We are grateful to the Neukom Foundation, the Linguistics Program, and the Native American Program at Dartmouth College for supporting the retreat. Likewise, many thanks to Michael Abramov and Oliver Adams for comments on this paper.

Speakers of minority languages frequently forgo literacy in their native languages to become competent in the dominant language of their nation state, which makes it difficult to obtain corpora for ASR models of minority languages. The first author was literate in Spanish and English long before she could read and write Chatino, which did not have a working alphabet until the author was forty years old.

Researchers of dominant languages, such as Spanish, do not face the same problem. For instance, Carlos Hernandez Mena, a computer scientist from UNAM who participated in the retreat related that he uses students doing their compulsory social service to transcribe and edit data that Carlos downloads from Youtube (Hernández Mena & Herrera 2017). This workflow has allowed him to acquire massive corpora of Mexican Spanish. Researchers of endangered languages do not have this same privilege.

In contrast, transcription of endangered languages is a long and roundabout process: a field linguist, usually with Western training, collaborates with a language consultant; they listen and discuss the sounds of the language together; the linguist makes notes and formulates questions; and the language consultant repeats those sounds over and over again.

Because of these obstacles, cross-disciplinary dialogue between linguists and computer scientists is a necessity. Computer scientists developing ASR models often do not understand the many difficulties that field linguists face: conflict-ridden locales, malaria outbreaks, or a severe lack of remaining language consultants. On the other hand, linguists are frequently unaware of the ASR technologies available to them.

High-tech industries do not develop ASR models for lesser-studied languages because they are not deemed profitable, while computational linguists often cite pressure to get university tenure as a reason to forgo research on minority languages. What has been written about tools for "low resource languages" in the literature of computer science focuses on the needs and desires of Western linguists, with little mention of native speakers and their role in the advancement of these tools. Collaborative conversations have not taken place in part because native speakers lack the influence, the funding, and the connections to convene researchers with a common vision. Most conversations about ASR take place in formal settings, such as conferences, workshops, and forums, and are not well attended by native speakers.

The need to automate the transcription of Chatino became more urgent for the first author when she began to transcribe audio recordings that Lynn Hou, an Assistant professor of Linguistics at the University of California Santa Barbara, made in San Juan Quiahije by working with families of deaf children. Lina is deaf and relies on transcription of any spoken language she encounters.

The author endeavored to provide Lina with careful annotations of the Chatino materials she collected so that she could have reliable data to analyze. The task was extremely time-consuming and laborious. The author found herself typing the same words over and over again, and she yearned to be able to automate the process. She began to ask linguists what it would take to automate the transcription of Chatino. She was told that most said ASR was not possible for minority languages because they lacked large corpora to feed the ASR models.

This question led her to collaborate with Damir and Malgorzata Cavar from Indiana University. By reading aloud numerous previously transcribed texts, they developed the first corpus of Chatino texts for ASR training (Cavar et al. 2016). This corpus has a creative commons license, meaning that anyone can download and use it. When the author became a Neukom Fellow at Dartmouth College, she set out to improve and expand on this corpus, while looking for ways to develop ASR for Chatino and other minority languages.

In another part of the world, linguist Alexis Michaud, who works with Yongning Na, a language of Southwest China, had similar goals: to automate the transcription process of the Na languages (Michaud et al. 2018). He designed his recordings so that they could eventually be used for ASR training. He began a successful collaboration with Oliver Adams, a computer scientist based out of the Melbourne University, who is now a Postdoctoral Fellow at Johns Hopkins University.

Along the way, Adams developed an open-source ASR toolkit called Persephone, which relies on neural networks, and quickly began to yield promising results in the transcription of Na. The tool was yielding a 20% error rate, and Michaud began to deploy Persephone into his linguistic workflow (Adams et al. 2018).They also found that Persephone could attain reasonable accuracy for a single speaker with as little as thirty minutes of data—an auspicious sign for endangered languages (Adams et al. 2018).

Next, Michaud and Adams wished to test the model on a comparable tonal language, and they found their way to the Chatino corpus in Global Open Resources and Information for Language and Linguistics Analysis. They invited me to evaluate some of Persephone's output (Adams et al. 2018), and to my great surprise, the system performed well for Chatino. At this time, Persephone is only accessible to computer scientists, requiring an interface overhaul to reach a broader audience. Seeing the results from the cross-comparison of Yongning Na and Chatino, motivated the first author to continue seeking collaborations with specialists in Natural language processing to continue advancing and improving ASR for Chatino and other indigenous languages of the Americas. This is how the idea of the retreat began.

**2.0 The retreat**
The William H. Neukom Foundation at Dartmouth College advocates for and funds interdisciplinary working groups to foster cross-disciplinary dialogue to advance scientific research. Often organized through "retreats," scholars with a common mission come together in a friendly and intimate place—usually the Upper Valley in New Hampshire and Vermont—to discuss solutions to a problem or a question that they have been pondering.

Daniel Rockmore, the dean of sciences at Dartmouth College and director of the Neukom Institute, encouraged the first author to host a retreat to discuss the development of ASR for endangered languages. With Rockmore's support, she proceeded to invite computer scientists, linguists, native speakers, and language activists to join me in Quechee, Vermont, where we would discuss ways to advance ASR for lesser-studied languages.[2] The event took place from July 12-14, 2018.

The twenty people who participated in the retreat came from John Hopkins, Carnegie Mellon, Yale University, University of North Texas, University of Texas at Austin, Universidad Autonoma de Mexico, and the Centro de Investigaciones y Estudios Superiores en Antropología, Mexico. The meeting struck a balance between engineers, linguists, native speakers, and language activists.

All of the participants stayed in a ski lodge called the Owl's Nest. The environment was low-key and relaxed. There were no PowerPoint presentations, just everyone gathered together in the living room. During catered meal breaks, participants organically broke into mixed groups and carried on the conversations of the day. We spent the first full day with introductions, sharing research interests, and setting an agenda for the weekend. Before dinner, we took an excursion paddling in canoes on the Connecticut River. The second day was marked with more detailed descriptions of everyone's research and concluded the day to a visit to Vermont Institute of Natural Science, a raptor education center and bird sanctuary.

On the second day, we shared and discussed what resources we had at our disposal for model training. Questions ranged from what languages participants had worked on and how many hours of recordings we had elicited, to more technical matters, such as recording format, file types (word, PDF, and ELAN),[3] to demographic information, including the number, age, and gender of speakers in the corpus.

Participants spoke or studied languages from six language families and four continents: from Australia: Nyulnyulan (Bardi) Pama-Nyungan (Djambarrpuyngu, Djapu); from Africa: Masso (Burkina Faso); from India: Tibeto-Burma (Manipuri and 17 others); and from the Americas: Otomanguean (Chatino, Otomi); Maya (Tzetsal, Tzotzil, and Mocho). Many of these languages are severely endangered. The Australian languages, for instance, have an average of four to five speakers, while one of the two varieties of Mocho Maya languages have one speaker each.

Both the linguists and the computer scientists were eager to learn about each other's workflows and the steps needed to complete a task. While most field linguists undergo an iterative process of data collection and documentation, computer scientists usually begin their research by reading academic papers and then replicating what they learn.

---

[2] ASREL retreat: https://sites.dartmouth.edu/neukom/asrel/)
[3] A software used by linguists for transcription, translation, and annotation.

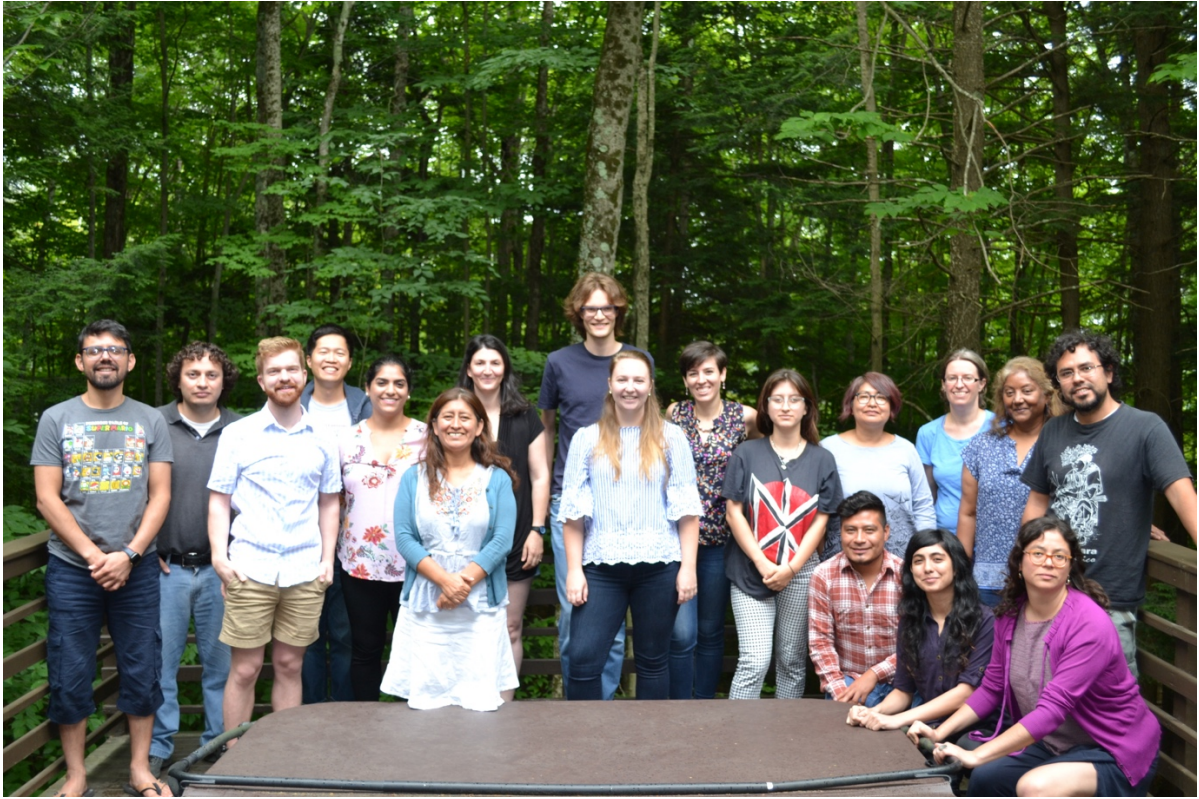

Though collaboration was our goal, we quickly learned that ASR models require substantial technical expertise to be used effectively. Persephone, for example, has preliminary support for ELAN files—which is great for linguists— and a web-API is under development so that Persephone might be used by a broader audience.

Moreover, we learned that there is no theoretical obstacle to creating an interface that would allow a linguist to upload speech and transcriptions for model training. It is largely a matter of having a professional software engineer develop the tool. Such automation has the potential to improve the rate of language documentation. Automating the transcription process yields three beneficial results. First, there is a potential for greater consistency in the transcription. Second, the researcher becomes less of a transcriber and more of an editor. Finally, automated transcriptions can provide fresh insight into the nature of the language under study.

We reached several milestones during our retreat. It was the first meeting of its kind to be convened by a native speaker of an endangered language. In the past, there has been little interest in ASR projects for minority languages. Native speakers have been historically denied the right to become literate in their languages, and as a result, those minority languages are under-represented and under-resourced in academia.

Native speakers need to be involved in conversations about ASR, as they bring community-oriented perspectives and accountability that are often overlooked or ignored by non-native speaker linguists and computer scientists. The event was a resounding success. It was productive and enjoyable. People left eager to continue the conversation on how improve and promote ASR for endangered languages.

**References**

Adams, Oliver & Cohn, Trevor & Neubig, Graham & Cruz, Hilaria & Bird, Steven & Michaud, Alexis. 2018. Evaluating phonemic transcription of low-resource tonal languages for language documentation. Proceedings of LREC 2018 (Language Resources and Evaluation Conference).. https://halshs.archives-ouvertes.fr/halshs-01709648.

Ćavar, Małgorzata E. &Cavar, Damir & Cruz, Hilaria. 2016. Endangered Language Documentation: Bootstrapping a Chatino Speech Corpus, Forced Aligner, ASR. *LREC*. 4004–4011.

Hernández Mena, Carlos Daniel & Herrera, Abel. 2017. CIEMPIESS (Corpus de Investigación en Español de México del Posgrado de Ingeniería Eléctrica y Servicio Social) Light. *Linguistic Data Consortium*. (https://catalog.ldc.upenn.edu/LDC2017S23).

Michaud, Alexis & Adams, Oliver & Cohn, Trevor A. & Neubig, Graham & Guillaume, Séverine. 2018. Integrating automatic transcription into the language documentation workflow: experiments with Na data and the Persephone toolkit. *Language Documentation and Conservation* 12. 393-429.

Mithun, Marianne. 1998. The significance of diversity in language endangerment and preservation. In L. Grenoble and L. Whaley (eds.), *Endangered languages: Current Issues and Future Prospects*, 163-191. Cambridge: Cambridge University Press.